\pdfoutput=1

\documentclass[11pt]{article}

\usepackage{acl}

\usepackage{latexsym, graphicx, float, amsmath, tabularx, subcaption}
\usepackage{booktabs}
\usepackage{threeparttable} 
\usepackage{times}
\usepackage{hyperref}
\usepackage{url}
\usepackage{multicol}
\usepackage{titlesec}
\usepackage{arydshln} 
\usepackage{inconsolata}
\usepackage{microtype}
\usepackage{amsfonts,amssymb}

\usepackage[T1]{fontenc}

\usepackage[utf8]{inputenc}
\usepackage{soul}

\title{Emphasising Structured Information: Integrating Abstract Meaning Representation into LLMs for Enhanced Open-Domain Dialogue Evaluation}
\newcommand*\samethanks[1][\value{footnote}]{\footnotemark[#1]}
\author{
Bohao Yang\textsuperscript{\rm 1}\thanks{\quad \small Equal contribution.},
Kun Zhao\textsuperscript{2}\samethanks, 
Dong Liu\textsuperscript{3}, 
Chen Tang\textsuperscript{\rm 4}\space, 
Liang Zhan\textsuperscript{2}\thanks{\quad \small Corresponding authors.},
Chenghua Lin\textsuperscript{1}\samethanks \\
\textsuperscript{1} The University of Manchester
\textsuperscript{2} University of the Pittsburgh \\
\textsuperscript{3} Tencent Timi Studio
\textsuperscript{4} University of Surrey\vspace{-0.5mm} \\
\texttt{
dougliu@tencent.com
}\vspace{-0.5mm} 
\texttt{
\{kun.zhao, liang.zhan\}@pitt.edu
}\vspace{0.4mm} \\
\texttt{
bohao.yang-2@postgrad.manchester.ac.uk
}\vspace{-0.1mm}
\texttt{
chenghua.lin@manchester.ac.uk,
}\vspace{-0.5mm}
}
\begin{document}
\maketitle
\begin{abstract}
Automatic open-domain dialogue evaluation has attracted increasing attention, yet remains challenging due to the complexity of assessing response appropriateness. Traditional evaluation metrics, typically trained with true positive and randomly selected negative responses, tend to assign higher scores to responses that share greater content similarity with contexts. However, adversarial negative responses, despite possessing high lexical overlap with contexts, can be semantically incongruous. Consequently, existing metrics struggle to effectively evaluate such responses, resulting in low correlations with human judgments.
While recent studies have demonstrated the effectiveness of Large Language Models (LLMs) for open-domain dialogue evaluation, they still face challenges in handling adversarial negative examples. We propose a novel evaluation framework that integrates Abstract Meaning Representation (AMR) enhanced domain-specific language models (SLMs) with LLMs. Our SLMs explicitly incorporate AMR graph information through a gating mechanism for enhanced semantic representation learning, while both SLM predictions and AMR knowledge are integrated into LLM prompts for robust evaluation.
Extensive experiments on open-domain dialogue evaluation tasks demonstrate the superiority of our method compared to state-of-the-art baselines. Our comprehensive ablation studies reveal that AMR graph information contributes substantially more to performance improvements. Our framework achieves strong correlations with human judgments across multiple datasets, establishing a new benchmark for dialogue evaluation. Our code and data are publicly available at \url{https://github.com/Bernard-Yang/SIMAMR}.
\end{abstract}

\section{Introduction}



Open-domain dialogue systems have garnered substantial attention owing to their broad applicability~\cite{zhao2023evaluating, Liu2023GEvalNE} across various domains, including personal medical assistance and biomedical telecommunications~\citep{Sai2020ImprovingDE, yang2024crafting}. Traditional evaluation approaches, such as $n$-gram-based metrics~\citep{Papineni2002BleuAM, Lin2004ROUGEAP, Banerjee2005METEORAA} and embedding-based metrics~\citep{Zhang2020BERTScoreET}, assess the semantic similarity between response candidates and gold references. These methods correlate poorly with human evaluation due to their limited capacity to incorporate conversational context~\citep{Liu2016HowNT}.
\begin{figure}[t]
\centering
\includegraphics[width=0.99\linewidth]{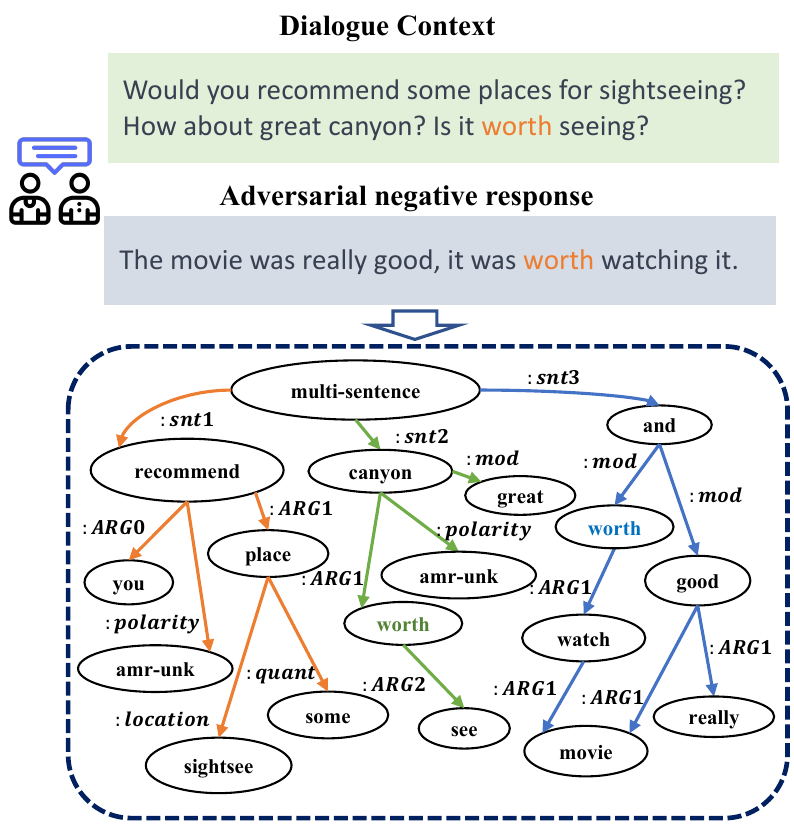}
\caption{AMR graphs for the conversational context and response. The semantic relationship of the word ``worth'' appearing in both context and response is captured through distinct colored representations in their respective AMR graphs.
}
\label{fig:amr}
\end{figure}

While recent advances in trainable evaluation frameworks~\citep{Lowe2017TowardsAA, Tao2018RUBERAU} have improved context-response relationship modeling, they face fundamental limitations stemming from their training . These models, typically trained with true positive and randomly sampled negative examples, tend to assess responses primarily through surface-level content similarity. Although some approaches have attempted to address this by incorporating adversarial examples~\citep{Sai2020ImprovingDE, Gupta2021SynthesizingAN}, they either require extensive pre-training on large-scale conversational corpora or demand adaptation to specific datasets, incurring substantial computational overhead. Moreover, their exclusive reliance on surface-form features compromises robustness when evaluating adversarial examples that deviate from the training distribution.
The vulnerability to adversarial attacks further compounds this challenge. \citet{Jin2019IsBR} demonstrated that even simple synonym substitutions can lead to misclassification in text analysis tasks. For instance, a positive review stating ``\textit{The characters, cast in impossibly \underline{contrived situations}, are \underline{totally} estranged from reality}'' would be misclassified as negative when minimally modified to ``\textit{The characters, cast in impossibly \underline{engineered circumstances}, are \underline{fully} estranged from reality}'', despite maintaining semantic equivalence.

Recent advances in Large Language Models (LLMs) have shown promise across a variety of tasks~\citep{ yang2023effective, Liu2023GEvalNE, yang2025does, Chiang2023CanLL}. However, these models still exhibit suboptimal performance when evaluating adversarial negative responses. To address these limitations, we propose integrating LLMs with domain-specific language models (SLMs) enhanced by Abstract Meaning Representation (AMR) graph information, specifically aimed at improving evaluation robustness for adversarial examples.
AMR graphs serve as powerful tools for capturing dialogue system states and providing complementary semantic knowledge~\citep{Bai2021SemanticRF, Bonial2020DialogueAMRAM}. Consider the following example: given the context ``\textit{Would you recommend some places for sightseeing? How about great canyon? Is it \underline{worth} seeing?}'', and an adversarial negative response ``\textit{The movie was really good, it was \underline{worth} watching it}'', existing metrics might erroneously classify this as positive due to lexical overlap. AMR graphs help address this by modeling semantic relationships between concepts (e.g., ``worth'' and ``canyon'') through explicit edge relations (e.g., ``:mod'' and ``:ARG1'').


Our approach introduces an AMR graph-enhanced SLM that effectively identifies adversarial negative examples in open-domain dialogue. The framework integrates both the SLM's predictions and AMR graph information into the LLM's prompt, creating a robust automatic evaluator that leverages domain-specific knowledge during inference.
The SLM architecture comprises two key components: sentence and graph encoders. The sentence encoder processes surface-form knowledge from conversational contexts and responses, while the graph encoder models AMR structural information, capturing both conceptual elements and their interrelations. These complementary representations are unified through a sophisticated gating mechanism and optimised via contrastive learning, encouraging alignment between textual and structural features for positive context-response pairs. The final evaluation integrates both the SLM's prediction score and AMR graph information into the LLM's prompt.

Comprehensive empirical evaluation across three public datasets demonstrates our model's superior performance compared to state-of-the-art baselines, including LLM-based methods. Our key contributions include:


Our contributions can be summarised as follows:

\begin{itemize}
\item A novel framework integrating AMR graph information into open-domain dialogue evaluation through a dual-representation approach that combines specialized SLMs with LLMs.

\item A comprehensive evaluation methodology across four distinct criteria (Naturalness, Coherence, Engagingness, and Groundedness), with detailed performance breakdowns demonstrating consistent improvements across all dimensions.

\item Extensive experimental results demonstrating substantial improvements over existing methods including reasoning-focused LLMs, with ablation studies revealing that AMR graph information contributes 7.4\% more to performance than SLM score alone.

\end{itemize}
\section{Related Work}

\noindent\textbf{Dialogue Evaluation Metrics.}~~Traditional $n$-gram-based metrics, including BLEU~\citep{Papineni2002BleuAM}, ROUGE~\citep{Lin2004ROUGEAP}, and METEOR~\citep{Banerjee2005METEORAA}, compute lexical overlap between response candidates and gold references. 
More sophisticated embedding-based metrics, such as Extrema~\citep{Forgues2014BootstrappingDS} and BERTScore~\citep{Zhang2020BERTScoreET}, first project responses and references into high-dimensional semantic spaces before calculating their similarity. 
However, both approaches have shown limited efficacy in evaluating open-domain dialogue systems~\citep{Liu2016HowNT}.

Regarding trainable metrics, RUBER~\citep{Tao2018RUBERAU} evaluates response quality by measuring semantic similarity between the generated response, dialogue context, and ground truth reference. \citet{Sai2020ImprovingDE} introduced DEB, which leverages a BERT model pre-trained on large-scale Reddit conversations. While effective, the computational cost of pre-training on extensive datasets makes this approach less practical. 
Similarly, Mask-and-fill~\citep{Gupta2021SynthesizingAN} employs a Speaker-Aware BERT architecture~\citep{Gu2020SpeakerAwareBF} to enhance dialogue understanding, though it requires dataset-specific adaptation before fine-tuning. 
\citet{Zhang2021MDDEvalSO} developed MDD-Eval for cross-domain dialogue evaluation, but this method necessitates human annotations and additional training data while failing to address adversarial negative examples.

\noindent\textbf{LLM-based Evaluators.}~~The emergence of Large Language Models (LLMs) has enabled new approaches to dialogue evaluation. \citet{Fu2023GPTScoreEA} developed GPTScore, leveraging pre-trained language models for multi-aspect, customizable evaluation without task-specific training. \citet{Wang2023IsCA} empirically validated the effectiveness of LLM-based evaluation approaches. \citet{Kocmi2023LargeLM} demonstrated the utility of GPT models in machine translation evaluation. \citet{Liu2023GEvalNE} introduced G-Eval, employing GPT-4 across multiple generation tasks including dialogue response, text summarization, data-to-text generation, and machine translation. \citet{chan2023chateval} proposed ChatEval, a multi-agent debate framework that surpasses single-LLM evaluators in performance. However, these LLM-based approaches have yet to be applied to evaluating adversarial negative responses incorporating non-textual domain knowledge.

\section{Methodology}
\begin{figure*}[ht]
\small
\centering 
\includegraphics[width=0.8\linewidth]{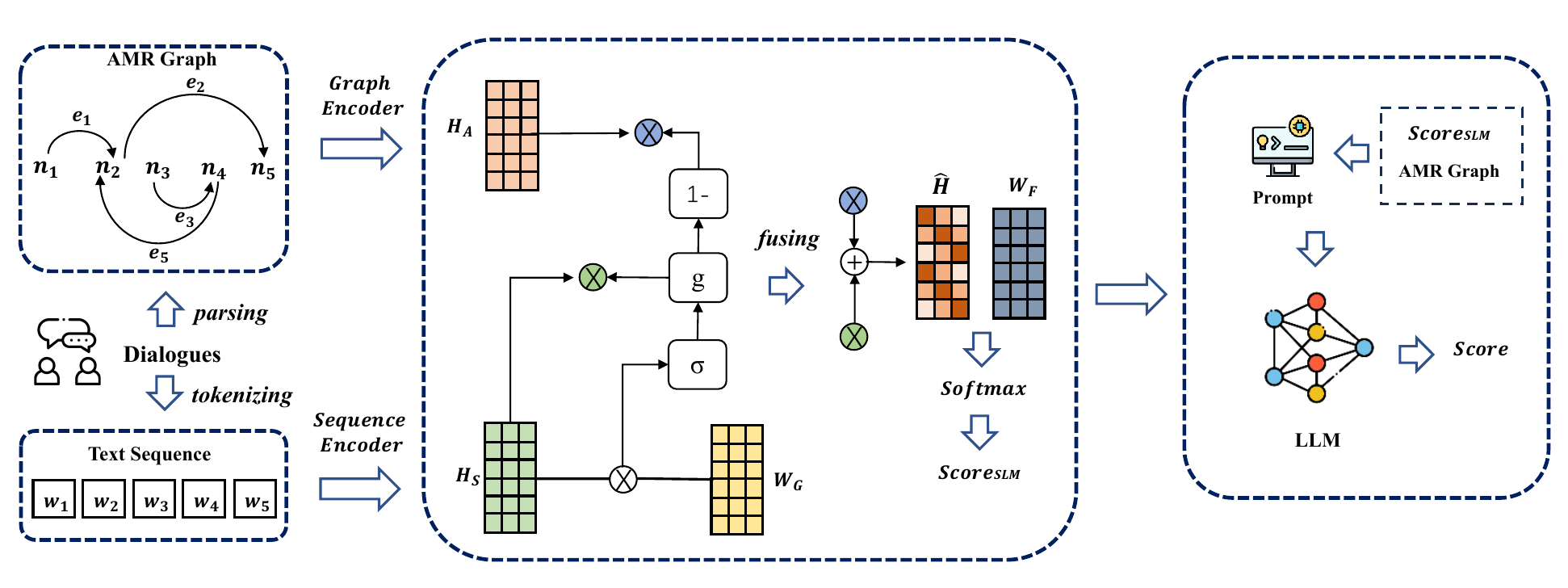}

\caption{The architecture of the proposed model. The left part is the SLM architecture, containing two encoders and the gate mechanism for encoding and fusing the sequence and AMR graph information of context-response pairs. The right part is the LLM where the prompt contains the prediction score of the SLM and AMR graph information.}
\label{fig:model}
\end{figure*}

\subsection{Task Description}
Our model operates on input tuples consisting of a dialogue context $\mathcal{C}$, a response $\mathcal{R}$, and their corresponding AMR graphs $\mathcal{G_C}$ and $\mathcal{G_R}$. The primary objective of the SLM component is to perform binary classification, predicting a label $\mathcal{Y} \in \{0,1\}$ for each response, where 0 and 1 denote negative and positive responses, respectively. 

The SLM generates a classification confidence score defined as:
\begin{equation}
\mathrm{Score_{SLM}} = P(\mathcal{Y} \mid \mathcal{C}, \mathcal{R}, \mathcal{G_C}, \mathcal{G_R})
\end{equation}

The derived confidence score, in conjunction with the semantic structural information encoded in AMR graphs $\mathcal{G_C}$ and $\mathcal{G_R}$, is incorporated into the LLM's prompt. This integration enables the LLM to leverage both statistical confidence and explicit semantic knowledge for more robust open-domain dialogue evaluation.
\subsection{Overall Architecture}

Figure~\ref{fig:model} illustrates the comprehensive architecture of our proposed framework, which seamlessly integrates SLM and LLM components. The SLM architecture incorporates a dual-encoder design: a sequence encoder for processing textual information and a graph encoder specialized in AMR graph representation learning. The complementary representations from these encoders are dynamically balanced through an adaptive gating mechanism, which modulates the information flow from both sources.

To optimise the alignment between textual and structural representations, particularly for positive response pairs, we employ a contrastive learning strategy during the training phase. This approach minimizes the representational distance between sentence and graph embeddings for semantically coherent pairs, while maintaining appropriate separation for negative examples. 

The final evaluation framework leverages both the SLM's classification confidence score $\mathrm{Score_{SLM}}$ and the structured AMR graph information, which are systematically integrated into the LLM's prompt through a carefully designed template. This multi-modal integration enables the LLM to synthesize both statistical and semantic evidence for more robust dialogue evaluation.

The complementary nature of SLM and LLM integration stems from their distinct capabilities: while the SLM excels at encoding structured graph information through specialized transformers, LLMs offer superior contextual reasoning but lack native graph processing abilities. As shown in our attention analysis in Appendix~\ref{app:att}, the SLM's graph encoder can identify semantic inconsistencies in adversarial examples that may be missed by text-only representations. By combining these approaches, our framework leverages both structured semantic knowledge and advanced reasoning capabilities.

\subsection{Sequence Encoder}

The sequence encoder employs a standard Transformer architecture~\citep{Vaswani2017AttentionIA} to process the input dialogue components. Given a dialogue context $\mathcal{C}_i = \left\{w_1, w_2, \ldots, w_\mathcal{C}\right\}$ and a response $\mathcal{R}_i = \left\{w_1, w_2, \ldots, w_\mathcal{R}\right\}$, where $w_i$ denotes the $i$-th token and $\mathcal{C}$, $\mathcal{R}$ represent respective sequence lengths, the encoder generates a sentence representation $\mathbf{H}_{S}$. The encoding process can be formally expressed as:

\begin{align}
\mathbf{H}_{S} &= \operatorname{SeqEncoder}(\mathcal{C,R}) \\
h_i &= \sum_{j=1}^{\mathcal{C+R}} \alpha_{i j}\left(W^H h_j\right) \\
\alpha_{i j} &= \operatorname{Attention}\left(h_i, h_j\right)
\end{align}

where $\mathbf{H}_{S} = \left\{h_1, h_2, \ldots, h_\mathcal{C+R}\right\}$ represents the sequence of hidden states and $W^H$ denotes the transformation matrix.

\subsection{Graph Encoder}
For modeling AMR graph structures, we utilise the Graph Transformer~\citep{Zhu2019ModelingGS}, an extension of the standard Transformer that specialises in graph-structured data. An AMR graph $\mathcal{G} = \langle\mathcal{V}, \mathcal{E}\rangle$ comprises nodes $\mathcal{V}$ and edges $\mathcal{E}$, where each edge $e \in \mathcal{E}$ is represented as a triple $\left\langle n_i, r_{i j}, n_j\right\rangle$ denoting the relation $r_{i j}$ between nodes $n_i$ and $n_j$. The graph encoding process is defined as:

\begin{align}
\mathbf{H}_{A} &= \operatorname{GraphEncoder}(\mathcal{V}, \mathcal{E}) \\
h'_i &= \sum_{j=1}^M \hat{\alpha}_{i j}\left(W^V h'_j + W^R \boldsymbol{r}_{i j}\right)
\end{align}

where $\mathbf{H}_{A} = \left\{h_1', h_2', \ldots, h_M'\right\}$ represents the graph embeddings, and $W^V$, $W^R$ are learnable transformation matrices.

The graph attention mechanism, which distinguishes the Graph Transformer from standard Transformers, is computed as:

\begin{align}
\hat{\alpha}_{i j} &= \frac{\exp \left(\hat{e}_{i j}\right)}{\sum_{m=1}^M \exp \left(\hat{e}_{i m}\right)} \nonumber \\ 
\hat{e}_{i j} &= \frac{\left(W^Q h'_i\right)^T\left(W^K h'_j + W^R \boldsymbol{r}_{i j}\right)}{\sqrt{d}}
\end{align}

where $W^Q$, $W^K$ are transformation matrices and $d$ is the dimensionality of the hidden states.

\subsection{Aggregation Gate}

To effectively combine the complementary information from both sequence and graph representations, we implement an adaptive gating mechanism. Given the sentence representation $\mathbf{H}_{S}$ and graph representation $\mathbf{H}_{A}$, the gate value $g_i$ is computed as:

\begin{align}
g_i &= \sigma\left(W^G \mathbf{H}_S + b_g\right) \\
\mathbf{\hat{H}} &= g_i \mathbf{H}_{S} + \left(1-g_i\right) \mathbf{H}_{A}
\end{align}

where $W^G$, $b_g$ are learnable parameters, and $\mathbf{\hat{H}}$ represents the final fused representation.

\subsection{Training objectives and Evaluation}

The fused representation $\mathbf{\hat{H}}$ is used to predict the classification probability for the context-response pair:

\begin{align}
\mathrm{Score_{SLM}} = \operatorname{softmax}\left(W^F \mathbf{\hat{H}} + b_f\right)
\end{align}

The training objective combines classification and contrastive learning:

\begin{align}
\mathcal{L} &= \mathcal{L}_{cls} + \mathcal{L}_C \label{eq:final loss} \\
\mathcal{L}_{cls} &= -\mathcal{\log} P(\mathcal{Y}=1 \mid \mathbf{\hat{H}})
\end{align}

The contrastive loss $\mathcal{L}_C$ facilitates alignment between sentence and graph representations:

\begin{align}
\mathcal{L}_C = -\frac{1}{N} 
\sum_{i=1}^N \frac{e^{\operatorname{sim}\left(\mathbf{H}_S^{+}, \mathbf{H}_A^{+}\right)}}
{\sum_j e^{\operatorname{sim}\left(\mathbf{H}_S^{-}, \mathbf{H}_A^{-}\right)}}
\label{eq:contrastive loss}
\end{align}

where $\mathbf{H}_S^{+}$, $\mathbf{H}_A^{+}$ denote positive pair representations and $\mathbf{H}_S^{-}$, $\mathbf{H}_A^{-}$ represent negative pairs.

The final evaluation score integrates the SLM prediction score $\mathrm{Score_{SLM}}$ and AMR graph information $\mathcal{G}$ through the LLM's prompt.

\begin{align}
\mathrm{Score} = \mathrm{LLMs}(\mathrm{Score_{SLM}}, \mathcal{G})
\end{align}

\begin{table*}
\centering
\begin{threeparttable}[ht]
\small
\resizebox{0.7\linewidth}{!}{
\begin{tabular}{lcc|cc}
             \toprule
& \multicolumn{2}{c|}{Standard Set} & \multicolumn{2}{c}{Adversarial Set}\\ 
Metrics & Pearson's $\rho$ & Spearman's $\tau$ & Pearson's $\rho$ & Spearman's $\tau$  \\ \midrule
BLEU-1 & 0.1841 (0.1620) & 0.1825 (0.1623) & 0.2064 (0.1321) & 0.2102 (0.9274)
\\
BLEU-2 & 0.1881 (0.1928) & 0.1772 (0.3928) & 0.1540 (0.3937) & 0.1969 (0.3921)
\\
BLEU-3 &0.1847 (0.4265) & 0.1835 (0.3521) & 0.1543 (0.4336) & 0.1973 (0.2292)
\\
BLEU-4 &0.1980 (0.2552) & 0.1787 (0.8398) & 0.1598 (0.6175) & 0.1844 (0.7698)
\\ 
ROUGE-1 &0.2183 (0.4698) & 0.2026 (0.7390) & 0.2305 (0.9120) & 0.2141 (0.4276)
\\
ROUGE-2 & 0.2055 (0.9153) & 0.1911 (0.1263) & 0.1516 (0.5291) & 0.1693 (0.5201)
\\
ROUGE-L &0.2183 (0.1028) & 0.2034 (0.1928) & 0.2377 (0.0183) & 0.2271 (0.1912)
\\ 
METEOR & 0.1804 (0.1018) & 0.1561 (0.1793) & 0.1342 (0.1123) & 0.1034 (0.5443)
\\ 
BERTScore & 0.2517 (0.3556) & 0.2658 (0.2369) & 0.2016 (0.3430) & 0.2230 (0.2561)\\
\midrule
DEB & 0.3236 (0.0630) & 0.2856 (0.2382) & 0.3492 (0.0622) & 0.3406 (0.8098) \\
USR & 0.2636 (0.0206) & 0.2482 (0.8432) & 0.2297 (0.0624) & 0.2760 (0.1892) \\
Mask-and-fill & 0.1904 (0.1732) & 0.2056 (0.0975) & 0.2604 (0.1320) & 0.2895 (0.0460)\\
MDD-Eval & 0.2813 (0.0610) & 0.2424 (0.8223) & 0.2982 (0.4162) & 0.2792 (0.0218) \\
\midrule
G-Eval (\texttt{GPT-3.5}) & 0.3418 (0.0106) & 0.3325 (0.0190) & 0.3294 (0.2327) & 0.3412 (0.2272) \\
QwQ-32B & 0.3915 (0.0123) & 0.3876 (0.0142) & 0.3783 (0.0224) & 0.3861 (0.0182) \\
Qwen2.5-7B & 0.3687 (0.0152) & 0.3702 (0.0134) & 0.3557 (0.0213) & 0.3674 (0.0198) \\
G-Eval (\texttt{GPT-4}) & 0.4321 (0.0001) & 0.4312 (0.0071) & 0.4298 (0.0225) & 0.4528 (0.0021) \\
LLM-Eval (\texttt{GPT-3.5}) & 0.3548 (0.0211) & 0.3723 (0.0190) & 0.3501 (0.3712) & 0.3421 (0.0762) \\
LLM-Eval (\texttt{GPT-4}) & 0.4315 (0.0206) & 0.4621 (0.0172) & 0.4691 (0.2355) & 0.4528 (0.5632) \\
\midrule
Ours(\texttt{w/o LLM}) & 0.3575 (0.0442) & 0.3646 (0.0347) & 0.3492 (0.0620) & 0.3545 (0.0215) \\
Ours (\texttt{GPT-3.5 w/o AMR}) & 0.4590 (0.0241) & 0.4592 (0.0539) & 0.4623 (0.2327) & 0.4745 (0.2342) \\
Ours (\texttt{GPT-3.5 w/o SLM})& 0.4782 (0.1242) & 0.4723 (0.0119) & 0.4898 (0.2237) & 0.4902 (0.0938) \\
Ours (\texttt{GPT-3.5}) & 0.4890 (0.0001) & 0.4873 (0.0019) & 0.4955 (0.1237) & 0.4920 (0.0462) \\

Ours (\texttt{GPT-4 w/o AMR}) & 0.5290 (0.2421) & 0.5392 (0.0129) & 0.5212 (0.2375) & 0.5522 (0.5632) \\
Ours (\texttt{GPT-4 w/o SLM})& 0.5426 (0.0106) & 0.5701 (0.0019) & 0.5521 (0.8375) & 0.5209 (0.9472) \\
Ours (\texttt{GPT-4}) & \textbf{0.5693 (0.0021)} & \textbf{0.5927 (0.0043)} 
& \textbf{0.5628 (0.0116)} & \textbf{0.5826 (0.0025)}  \\

\bottomrule

\end{tabular}
}
\end{threeparttable}
\caption{Pearson and Spearman correlations with human judgments on the DailyDialog++ dataset. The number figures in parentheses are p-values.}

\label{tab:correlation1}
\end{table*}
\section{Experiments}
\subsection{Dataset}
We conduct experiments on three widely-recognised open-domain dialogue datasets: \textbf{DailyDialog++}~\citep{Sai2020ImprovingDE}, \textbf{PersonaChat}~\citep{zhang-etal-2018-personalizing}, and \textbf{TopicalChat}~\cite{gopalakrishnan2019topical}.
DailyDialog++ is particularly noteworthy as it is the sole publicly available dataset containing human-crafted adversarial negative responses. 
Each context is paired with three types responses: five positive responses, five random negative responses, and five adversarial negative responses.

For PersonaChat and TopicalChat, which lack human-created adversarial responses in their original forms, we utilise the augmented datasets from~\citep{zhao2024slide}. These enhanced datasets feature 2,000 conversational contexts, each accompanied by five positive responses and adversarial negative counterparts.

\subsection{Experimental Settings}

The preprocessing of AMR graph structures involves multiple stages. Initially, we employ the \textit{amrlib} library~\citep{Cai2020AMRPV} to transform each context-response pair into its corresponding AMR graph representation. Following the methodology outlined in~\citep{Song2020StructuralIP}, we subsequently process these graphs using the AMR simplifier~\citep{Konstas2017NeuralAS}. This procedure include the error-checking and therefore yields refined and accurate AMR graphs.
For the LLM component, we utilise \texttt{GPT-3.5-turbo} and \texttt{GPT-4-1106}. 
The SLM is trained on the DailyDialog++ dataset, which comprises 9,259 dialogue contexts in the training set, 1,028 in the validation set, and 1,142 in the test set. 

\begin{table*}
\centering
\begin{threeparttable}[b]
\small
\resizebox{0.7\linewidth}{!}{
\begin{tabular}{lcc|cc}
             \toprule
& \multicolumn{2}{c|}{Standard Set} & \multicolumn{2}{c}{Adversarial Set}\\ 
Metrics & Pearson's $\rho$ & Spearman's $\tau$ & Pearson's $\rho$ & Spearman's $\tau$  \\ \midrule
BLEU-1 & 0.2063 (0.9228) & 0.2152 (0.6538) & 0.1764 (0.2243)  & 0.1663 (0.0335)\\
BLEU-2 & 0.1951 (0.7401) & 0.1823 (0.1361) & 0.1405 (0.3621)  & 0.1619 (0.1422) \\
BLEU-3 & 0.1680 (0.3465) & 0.1941 (0.8264) & 0.1375 (0.2103)  & 0.1676 (0.3456) \\
BLEU-4 & 0.2002 (0.2836) & 0.1930 (0.1712) &  0.1253 (0.0924) & 0.1543 (0.8927) \\
ROUGE-1 & 0.2130 (0.4942) & 0.2159 (0.3892) & 0.2075 (0.5918) & 0.2198 (0.1984) \\
ROUGE-2 & 0.2016 (0.0183) & 0.2023 (0.9172) & 0.1832 (0.1830) & 0.2073 (0.1983) \\
ROUGE-L & 0.2103 (0.9028) & 0.2034 (0.9283) & 0.2027 (0.9278) & 0.2236 (0.9183) \\
METEOR & 0.1997 (0.0183) & 0.1768 (0.0918) & 0.1439 (0.9214)  & 0.1705 (0.4028) \\
BERTScore & 0.2865 (0.2357) & 0.2721 (0.2568) & 0.2254 (0.5914) &  0.2643 (0.6019) \\
\midrule

DEB & 0.3653 (0.0241) & 0.3434 (0.8346) & 0.3512 (0.0301) & 0.3706 (0.8398) \\
USR & 0.3466 (0.0392) & 0.3456 (0.1343) & 0.3681 (0.0462) & 0.3859 (0.1846) \\
MDD-Eval & 0.3481 (0.0619) & 0.3410 (0.1802) &  0.3735 (0.1503) &  0.3601 (0.9348) \\
Mask-and-fill & 0.3093 (0.1812) & 0.3105 (0.8013) & 0.3764 (0.3153) & 0.3613 (0.2203) \\
\midrule
G-Eval (\texttt{GPT-3.5}) & 0.4891 (0.0923) & 0.4874 (0.0122) & 0.4551 (0.0410) & 0.4610 (0.0512) \\
QwQ-32B & 0.5027 (0.0124) & 0.5006 (0.0132) & 0.4778 (0.0215) & 0.4827 (0.0164) \\
Qwen2.5-7B & 0.4792 (0.0146) & 0.4734 (0.0129) & 0.4623 (0.0218) & 0.4707 (0.0173) \\
G-Eval (\texttt{GPT-4})  & 0.5241 (0.0131) & 0.5313 (0.0424) & 0.5123 (0.0112) & 0.5513 (0.0253) \\
LLM-Eval (\texttt{GPT-3.5}) & 0.4648 (0.1821) & 0.4573 (0.9181) & 0.4450 (0.7163) & 0.4614 (0.7817) \\
LLM-Eval (\texttt{GPT-4}) & 0.5321 (0.8127) & 0.5392 (0.7161) & 0.5269 (0.9221) & 0.5258 (0.9271) \\

\midrule

Ours(\texttt{w/o LLM}) & 0.3668 (0.0044) & 0.3784 (0.0037) & 0.3954 (0.0060) & 0.3911 (0.0055) \\
Ours (\texttt{GPT-3.5 w/o AMR}) & 0.5007 (0.0032) & 0.4998 (0.0008) & 0.5011 (0.0237) & 0.5105 (0.0047) \\
Ours (\texttt{GPT-3.5 w/o SLM})& 0.5118 (0.0024) & 0.5068 (0.0038) & 0.5199 (0.0007) & 0.5187 (0.0005) \\
Ours(\texttt{GPT-3.5}) & 0.5517 (0.0044) & 0.5209 (0.0002) & 0.5204 (0.0053) & 0.5225 (0.0057) \\

Ours (\texttt{GPT-4 w/o AMR}) & 0.6199 (0.0001) & 0.6127 (0.0004) & 0.6178 (0.0017) & 0.6004 (0.0028) \\
Ours (\texttt{GPT-4 w/o SLM})& 0.6267 (0.0021) & 0.6299 (0.0003) & 0.6245 (0.0047) & 0.6309 (0.0145) \\
Ours (\texttt{GPT-4}) & \textbf{0.6598 (0.0021)} & \textbf{0.6604 (0.0023)} & \textbf{0.6526 (0.0013)} & \textbf{0.6612 (0.0046)} \\

\bottomrule

\end{tabular}
}
\end{threeparttable}
\caption{Pearson and Spearman correlations with human judgments on the PersonaChat dataset.
}

\label{tab:correlation1.5}
\end{table*}

\begin{table*}
\centering
\begin{threeparttable}[b]
\small
\resizebox{0.7\linewidth}{!}{
\begin{tabular}{lcc|cc}
             \toprule
& \multicolumn{2}{c|}{Standard Set} & \multicolumn{2}{c}{Adversarial Set}\\ 
Metrics & Pearson's $\rho$ & Spearman's $\tau$ & Pearson's $\rho$ & Spearman's $\tau$  \\ \midrule
BLEU-1 & 0.2102 (0.2993) & 0.1982 (0.8628) & 0.1444 (0.0203)   & 0.1553 (0.0032)
 \\
BLEU-2 & 0.1721 (0.7761) & 0.1772 (0.3132) & 0.1295 (0.4321)  & 0.1439 (0.5402) \\
BLEU-3 & 0.1577(0.1357) & 0.1642 (0.1854) & 0.1225 (0.0203)  & 0.1328 (0.0341) \\
BLEU-4 & 0.1482 (0.2901) & 0.1503(0.1709) &  0.1323 (0.0203) & 0.1228 (0.3265) \\
ROUGE-1 & 0.2050 (0.4808) & 0.2144 (0.0371) & 0.1752 (0.2839) & 0.1788 (0.6052) \\
ROUGE-2 & 0.2005 (0.0956) & 0.2027 (0.1231) & 0.1835 (0.4462) & 0.2028 (0.2302) \\
ROUGE-L & 0.2197 (0.4980) & 0.2011 (0.3924) & 0.1908 (0.2993) & 0.2335 (0.7158) \\
METEOR & 0.1857 (0.1314) & 0.1576 (0.4371) & 0.1518 (0.8903)  & 0.1685 (0.4094) \\
BERTScore & 0.2555 (0.6227) & 0.2542 (0.9268) & 0.2194 (0.1936) &  0.2558 (0.2032) \\
\midrule

DEB & 0.3255 (0.0152) & 0.3306 (0.0470) & 0.3419 (0.0158) & 0.3668 (0.0812) \\
USR & 0.3466 (0.0045) & 0.3428 (0.1257) & 0.3338 (0.0478) & 0.1706 (0.0462) \\
MDD-Eval & 0.3277 (0.0245) & 0.3398 (0.2784) &  0.3869 (0.3478) &  0.3557 (0.0254) \\
Mask-and-fill & 0.2998 (0.0458) & 0.3052 (0.0025) & 0.3668 (0.1069) & 0.3627 (0.0044) \\
\midrule
G-Eval (\texttt{GPT-3.5}) & 0.4995 (0.0025) & 0.4754 (0.0011) & 0.4774 (0.0069) & 0.4688 (0.0098) \\
QwQ-32B & 0.5092 (0.0118) & 0.4836 (0.0125) & 0.4887 (0.0208) & 0.4824 (0.0152) \\
Qwen2.5-7B & 0.4927 (0.0137) & 0.4703 (0.0114) & 0.4702 (0.0211) & 0.4678 (0.0167) \\
G-Eval (\texttt{GPT-4})  & 0.5314 (0.0028) & 0.5055 (0.0015) & 0.4995 (0.0057) & 0.5022 (0.0064) \\
LLM-Eval (\texttt{GPT-3.5}) & 0.4837 (0.0001) & 0.4798 (0.0004) & 0.4512 (0.0007) & 0.4799 (0.0004) \\
LLM-Eval (\texttt{GPT-4}) & 0.5008 (0.0022) & 0.5096 (0.0036) & 0.5178 (0.0019) & 0.5257 (0.0007) \\

\midrule

Ours(\texttt{w/o LLM}) & 0.3602 (0.0011) & 0.3599 (0.0004) & 0.3611 (0.0017) & 0.3587 (0.0023) \\
Ours (\texttt{GPT-3.5 w/o AMR}) & 0.5022 (0.0001) & 0.5120 (0.0009) & 0.5118 (0.0025) & 0.5099 (0.0002) \\
Ours (\texttt{GPT-3.5 w/o SLM})& 0.5172 (0.0025) & 0.5099 (0.0065) & 0.5112 (0.0004) & 0.5101 (0.0051) \\
Ours(\texttt{GPT-3.5}) & 0.5200 (0.0051) & 0.5115 (0.0007) & 0.5127 (0.0057) & 0.5110 (0.0001) \\

Ours (\texttt{GPT-4 w/o AMR}) & 0.6274 (0.0001) & 0.6266 (0.0019) & 0.6198 (0.1237) & 0.5207 (0.0272) \\
Ours (\texttt{GPT-4 w/o SLM})& 0.6470 (0.0021) & 0.6482 (0.0031) & 0.6398 (0.0004) & 0.6402 (0.0054) \\
Ours (\texttt{GPT-4}) & \textbf{0.6641 (0.0002)} & \textbf{0.6603 (0.0002)} & \textbf{0.6598 (0.0007)} & \textbf{0.6674 (0.0003)} \\

\bottomrule

\end{tabular}
}
\end{threeparttable}
\caption{Pearson and Spearman correlations with human judgments on the TopicalChat dataset.}

\label{tab:correlation2}
\end{table*}
\subsection{Baselines}
For the word-overlap and embedding-based metrics, we select widely used ones in generative dialogue systems, including BLEU~\citep{Papineni2002BleuAM}, ROUGE~\citep{Lin2004ROUGEAP}, METEOR~\citep{Banerjee2005METEORAA}, and BERTScore~\citep{Zhang2020BERTScoreET}. 
For the learning-based metrics, We compare our method with DEB~\cite{Sai2020ImprovingDE}, USR~\cite{mehri-eskenazi-2020-usr}, Mask-and-fill~\citep{Gupta2021SynthesizingAN}, and MDD-Eval~\citep{Zhang2021MDDEvalSO}. Additionally, we select G-Eval~\citep{Liu2023GEvalNE}, QWQ-32B~\cite{qwq32b}, Qwen2.5-7B~\cite{qwen2.5}, and LLM-Eval~\cite{lin-chen-2023-llm} as the LLM-based metrics. For Qwen2.5-7B, we fine-tuned it on 12,000 both text and AMR structured dialogue examples from all three datasets, ensuring no overlap with evaluation sets.

\subsection{Evaluation Set and Human Annotation}
To rigorously assess our proposed metric, we establish a comprehensive evaluation protocol comprising two distinct sets: a \textit{Standard Set} and an \textit{Adversarial Set}. 

\noindent\textbf{Dataset Construction}~~The Standard Set encompasses positive and random negative responses, with 400 context-response pairs sourced from each of DailyDialog++, PersonaChat, and TopicalChat datasets, totalling 1,200 samples. The random negative responses are selected from different dialogue turns to ensure contextual diversity. The Adversarial Set, designed to evaluate robustness against challenging examples, contains an additional 400 context-response pairs per dataset, featuring positive and adversarial negative responses. In aggregate, our evaluation corpus comprises 2,400 context-response pairs.


\noindent\textbf{Correlation Computation}~~For reporting our experimental results, we compute correlation between automated scores and human judgments separately for each of the four criteria (Naturalness, Coherence, Engagingness, and Groundedness). The reported values in Tables~\ref{tab:correlation1}-\ref{tab:correlation2} represent the average correlations across all four dimensions. This approach follows standard practices in dialogue evaluation research~\citep{mehri-eskenazi-2020-usr}. A detailed breakdown of performance across individual criteria is provided in Appendix~\ref{app:breakdown}.

\noindent\textbf{Human Annotation}~~
Three qualified human evaluators, each holding at least a master's degree in Computer Science and demonstrating full professional English proficiency, independently rated each context-response pair. Assessments were conducted using a 5-point Likert scale, where higher scores indicate superior quality. The final human annotation score for each aspect was derived by averaging across all evaluators.
To ensure annotation reliability, we computed the Inner-Annotator Agreement (IAA) using Cohen's Kappa coefficient~\citep{cohen1960coefficient}. The achieved average IAA score of 0.64 between annotator pairs indicates substantial agreement (0.6-0.8), validating the robustness of our human evaluation framework.

\section{Results}

\subsection{Evaluation Performance on Standard Set}

We evaluate our model against the baselines by analysing the correlation between automated evaluation scores and human judgements across three datasets. The results presented in Table~\ref{tab:correlation1} to Table~\ref{tab:correlation2} reveal that $n$-gram and embedding-based baselines, which compute word overlap or semantic similarity between gold references and responses, demonstrate weak positive correlations with human annotations across two datasets. Amongst the $n$-gram baselines, ROUGE-L exhibits the strongest correlation. 
The embedding-based approach, BERTScore, whilst outperforming the $n$-gram baselines, still achieves suboptimal performance when compared with more sophisticated metrics. Learning-based metrics, which consider the contextual relationship between dialogue pairs, demonstrate superior overall performance. Specifically, Mask-and-fill and USR achieve better correlations than $n$-gram baselines, whilst DEB and MDD-Eval secure higher correlations among these approaches.

Regarding LLM-based methods, G-Eval and LLM-Eval demonstrate strong performance across all three datasets. We also evaluated reasoning-focused LLMs including QwQ-32B (via direct prompting without AMR) and Qwen2.5-7B (fine-tuned with structured data for 5 epochs). These models perform slightly better than GPT-3.5 across all datasets. Similarly, the fine-tuned Qwen2.5-7B (0.3687/0.3702) outperforms GPT-3.5, demonstrating the potential of specialized reasoning models.

Our method in its basic configuration (Ours w/o LLM) achieves moderately positive correlations across the three datasets (less than 0.4). However, when integrating SLM with LLM, our approach achieves the highest overall performance on both Pearson and Spearman correlations across all datasets. Notably, our GPT-4 variant exhibits superior performance compared to all baselines, including the reasoning-focused LLMs. Through ablation studies examining the effectiveness of SLM and AMR graphs, we observe that Ours (w/o SLM) outperforms Ours (w/o AMR), which combines only LLM and SLM components, thereby validating the effectiveness of incorporating AMR graphs in open-domain dialogue evaluation.

\subsection{Evaluation Performance on Adversarial Set}

To evaluate our method's capability in evaluating adversarial negative examples, we conduct comparative analyses against baseline approaches on the adversarial set. Tables~\ref{tab:correlation1} to~\ref{tab:correlation2} present the correlation results between automated metrics and human judgements.
The $n$-gram and embedding-based metrics exhibit weakly positive correlations with human judgements, primarily due to their inherent limitation of solely comparing gold references with response candidates, without considering the contextual relationships that characterise adversarial examples. 
Regarding learning-based approaches, USR demonstrates limited robustness against adversarial negative examples, showing only weak positive correlations with human judgements. In contrast, MDD-Eval, Mask-and-fill, and DEB achieve notably stronger performance across both Pearson and Spearman correlations. 

LLM-based methods establish themselves as the strongest baseline approaches, with reasoning-focused models like QwQ-32B and fine-tuned Qwen2.5-7B showing improved performance over standard GPT-3.5. 
However, despite these improvements, these reasoning-focused LLMs still fall short of our full approach, suggesting that explicit semantic structure through AMR graphs provides complementary information that enhances evaluation capabilities beyond what these models can derive from text alone.

Our proposed metric consistently surpasses all baseline approaches across both correlation metrics. Specifically, Ours(GPT-4) achieves strong correlations on the adversarial set, significantly outperforming the strongest baseline G-Eval. Similar improvements are observed in Spearman correlations across the three datasets. The ablation analysis further substantiates the benefits of our integrated approach: Ours(w/o AMR) shows notably lower correlations, demonstrating that the incorporation of AMR graph information significantly enhances the model's ability to evaluate adversarial examples. These results comprehensively validate the effectiveness of integrating AMR graph-enhanced SLM with LLMs for robust open-domain dialogue evaluation.



\subsection{Ablation Study}

We evaluate our SLM's classification performance on the DailyDialog++ testset. As shown in Table~\ref{table:Accu}, our SLM surpasses all baselines and demonstrating the effectiveness of incorporating AMR graph information.
Ablation studies reveal that removing either the Graph Transformer or Sentence Transformer components of SLM leads to decreased performance, with the Graph Transformer alone performing marginally better than the Sentence Transformer. While removing the contrastive learning (CL) or gating mechanism (GM) shows minimal impact, the removal of AMR information results in the most significant performance drop, highlighting its crucial role in dialogue evaluation.

When comparing Ours (w/o AMR) and Ours (w/o SLM) variants, we observe that removing AMR graph information leads to a more significant performance drop than removing the SLM score, confirming that the structured semantic knowledge encoded in AMR graphs contributes more to performance improvements. However, the full model combining both components achieves the best results, demonstrating that the SLM's specialized architecture for processing graph information and the LLM's reasoning capabilities operate synergistically rather than redundantly.

\section{Conclusion}

In this paper, we presents a novel evaluation framework for open-domain dialogue systems that integrates AMR graph-enhanced SLMs with LLMs. Comprehensive experimental results across multiple datasets demonstrate that our method consistently outperforms existing approaches, including state-of-the-art LLM-based methods, in the challenging task of open-domain dialogue evaluation. 


\section*{Ethics Statement}
Our proposed evaluation metric enhances the assessment of open-domain dialogue systems through AMR integration and contrastive learning. While the framework effectively addresses the one-to-many nature of dialogue evaluation, it may occasionally assign high scores to inappropriate responses. We recommend careful screening of training data and implementation of content filters before deployment.



\section*{Limitations}
Despite demonstrating robust performance, our method primarily focuses on semantic dependencies between context and response. Following \citet{Howcroft2020TwentyYO}, we acknowledge that human evaluation encompasses multiple attributes beyond semantic relationships. Future work should explore disentangling these various attributes to enhance model interpretability and evaluation comprehensiveness.
\bibliography{custom, zotero}

\newpage
\appendix
\section{More Experimental Results and Analysis}

\subsection{Case Study}
To demonstrate the effectiveness of AMR graphs in identifying adversarial negative responses, we present several illustrative examples in Table~\ref{tab:case study1}. These cases highlight instances where responses were incorrectly classified as ``positive'' without AMR graph information, but were accurately identified as ``negative'' when incorporating semantic structural knowledge from AMR graphs. This analysis underscores the crucial role of AMR-derived semantic information in enhancing the model's discriminative capability for challenging adversarial examples.

\begin{table}[h]
\small
\begin{tabular}{llll}
\toprule

\multicolumn{1}{l}{\textbf{Context:}}& \multicolumn{3}{l}{\begin{tabular}{p{5cm}}
Hi kevin, how was your year at college? It was great! How was your year? It was good. Do you have a \textbf{girlfriend} at \textbf{school}? \end{tabular}}\\ \hline 
\multicolumn{1}{l}{\textbf{Response:}} & \multicolumn{3}{l}{\begin{tabular}{p{5cm}}
Are you still in touch with any of your old \textbf{school friends}?\end{tabular}}\\  \hline

\multicolumn{1}{l}{\textbf{Context:}} & \multicolumn{3}{l}{\begin{tabular}{p{5cm}}Would you recommend some \textbf{places} for sightseeing? How about great canyon? Is it worth seeing?
\end{tabular}} \\ \hline
\multicolumn{1}{l}{\textbf{Response:}} & \multicolumn{3}{l}{\begin{tabular}{p{5cm}}Singapore is reportedly a very exciting \textbf{place} to live.\end{tabular}} \\ \hline

\multicolumn{1}{l}{\textbf{Context:}}& \multicolumn{3}{l}{\begin{tabular}{p{5cm}}I need change for the \textbf{machines}? You need to put \textbf{50} cents into the washer \textbf{machine} and a dollar into the dryer. So what do I need to do?  \end{tabular}}\\\hline 
\multicolumn{1}{l}{\textbf{Response:}} & \multicolumn{3}{l}{\begin{tabular}{p{5cm}}In our factory, there are \textbf{50} electrical \textbf{machines}.\end{tabular}}                           \\ 
\bottomrule
\end{tabular}

\caption{Samples of context-response pairs. The bold words represent the overlapping words.}
\label{tab:case study1}
\end{table}

\subsection{Attention Visualisation Analysis}
\label{app:att}
\begin{figure*}[t]
\centering 
    \begin{subfigure}{\textwidth}
        \centering 
        \includegraphics[width=0.95\linewidth]{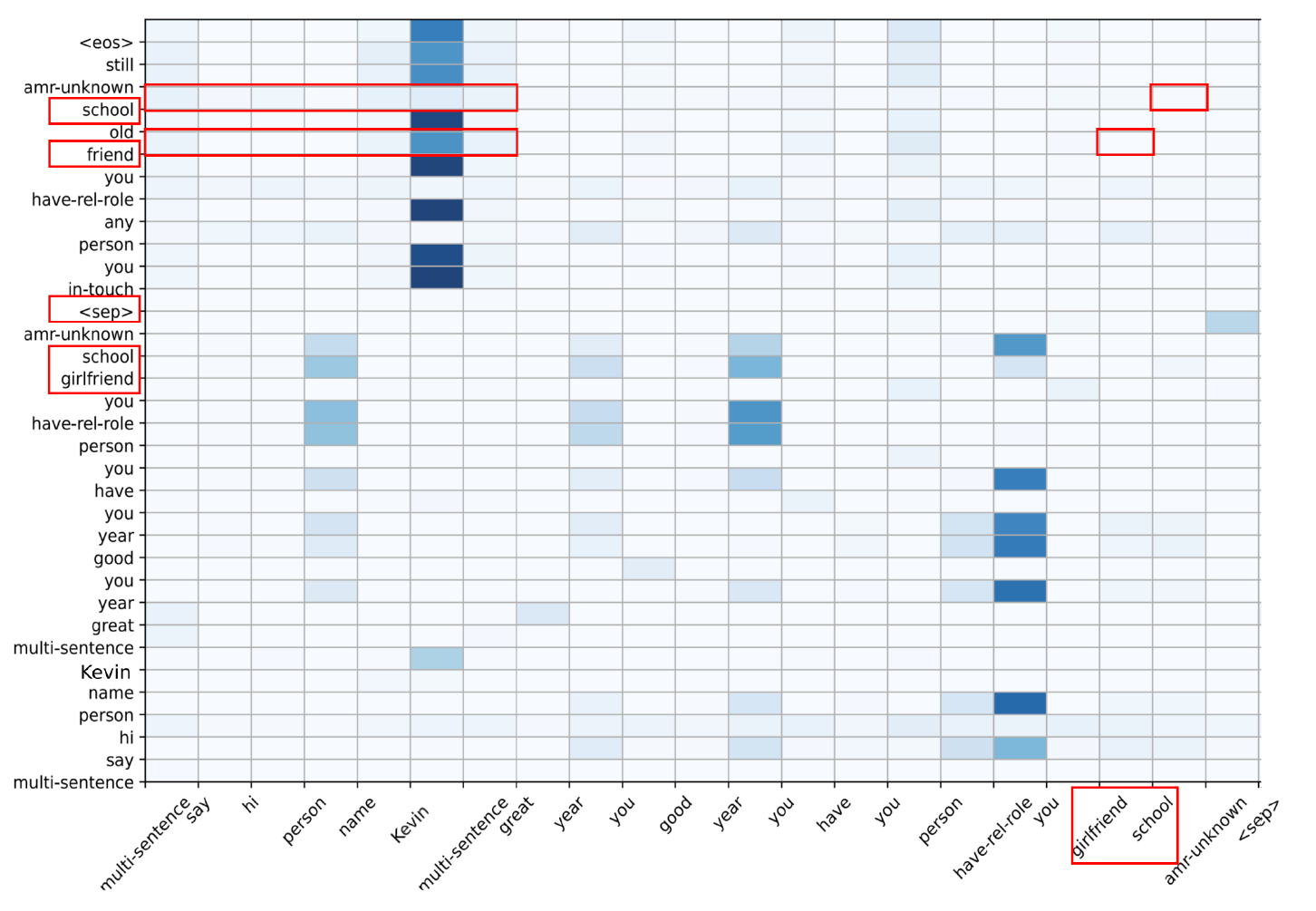}
    \end{subfigure}
    
    \begin{subfigure}{\textwidth}
        \centering 
        \includegraphics[width=0.95\linewidth]{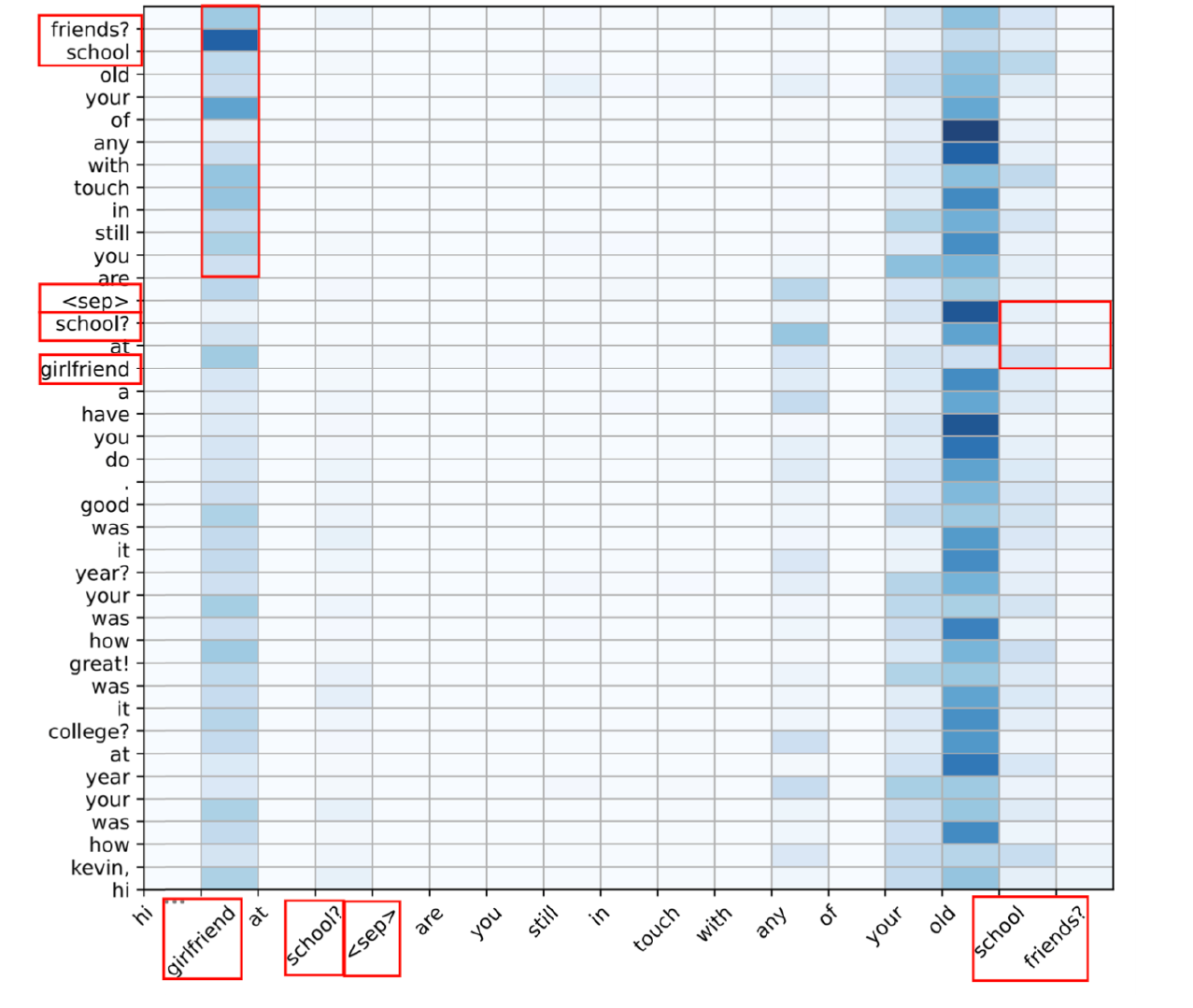}
    \end{subfigure}

\caption{Attention pattern visualisation for context-response analysis. Top: Graph Transformer attention heatmap showing semantic-aware attention distribution. Bottom: Sentence Transformer attention heatmap highlighting lexical-level attention patterns. Overlapping tokens between context and response (\textit{friends} and \textit{school}) demonstrate distinct attention behaviours in the two encoders.}
\label{fig:case}
\end{figure*}

We analyse the attention patterns of both Sentence and Graph Transformers of the SLM through visualisation of their attention heatmaps for the context-response pair shown in Figure~\ref{fig:case}.

The Sentence Transformer exhibits strong attention weights between overlapping tokens in context and response. As illustrated in Figure~\ref{fig:case} (top), tokens such as ``school'' and ``friend'' in the response show high attention scores with their counterparts ``school'' and ``girlfriend'' in the context.
In contrast, the Graph Transformer, which incorporates entity relationships through AMR structures, demonstrates different attention patterns. Figure~\ref{fig:case} (bottom) shows that these lexically similar tokens receive lower attention weights, indicating the model's ability to capture semantic differences beyond surface-level similarities.

\begin{table}[]
\small
\centering
\begin{tabular}{lc}
\toprule
\multicolumn{1}{l}{Model} & Accuracy \\
\midrule

BERT Regressor & 75.92 \\
RUBER  & 76.50 \\ 
DEB  & 82.04 \\
\text{Mask-and-fill} & 85.27 \\
\midrule
\begin{tabular}[c]{@{}l@{}@{}}\textbf{Ours (SLM)}\\   
Ours (- w/o GM) \\   
Ours (- w/o CL) \\
Ours (- w/o GM, CL)
\end{tabular} &
\begin{tabular}[c]{@{}l@{}@{}}
\textbf{86.81} \\
86.22\\ 
86.46\\ 
85.64\\ 
\end{tabular} \\ 
Graph Transformer & 84.73\\ 
Sentence Transformer & 83.81 \\
\bottomrule
\end{tabular}
\caption{Ablation study on Dailydialog++ dataset.}
\label{table:Accu}
\end{table}

\section{Prompt Templates}
\label{app:prompt}

\subsection{Prompt for Engagingness evaluation}

\texttt{Rate the dialogue response.\\
Use the prediction probability from the SLMs and AMR graphs of the conversation pair to aid your judgment. \\
Note: Please take the time to fully read and understand the dialogue response. \\
How dull/interest is the text of the dialogue response? (on a scale of 1-5, with 1 being the lowest)\\
Input:\\
Conversation Context:\\
Response:\\
AMR Graph:\\
SLM score: \\\\
Evaluation Form (Score ONLY):\\
Engagingness: \\
}

\subsection{Prompt for Naturalness evaluation}
\texttt{Rate the dialogue response.\\
Use the prediction probability from the SLMs and AMR graphs of the conversation pair to aid your judgment. \\
Note: Please take the time to fully read and understand the dialogue response. \\
To what extent the response is naturally written (on a scale of 1-5, with 1 being the lowest)\\
Input:\\
Conversation Context:\\
Response:\\
AMR Graph:\\
SLM score: \\\\
Evaluation Form (Score ONLY):\\
Naturalness: \\
}

\subsection{Prompt for Coherence evaluation}

\texttt{Rate the dialogue response.\\
Use the prediction probability from the SLMs and AMR graphs of the conversation pair to aid your judgment. \\
Note: Please take the time to fully read and understand the dialogue response. \\
To what extent the response is well-structured, logical, and meaningful (on a scale of 1-5, with 1 being the lowest)\\
Input:\\
Conversation Context:\\
Response:\\
AMR Graph:\\
SLM score: \\\\
Evaluation Form (Score ONLY):\\
Coherence: \\
}

\subsection{Prompt for Groundedness evaluation}

\texttt{Rate the dialogue response.\\
Use the prediction probability from the SLMs and AMR graphs of the conversation pair to aid your judgment. \\
Note: Please take the time to fully read and understand the dialogue response. \\
To what extent the response is grounded in facts present in the context (on a scale of 1-5, with 1 being the lowest)\\
Input:\\
Conversation Context:\\
Response:\\
AMR Graph:\\
SLM score: \\\\
Evaluation Form (Score ONLY):\\
Groundedness: \\
}

\begin{table*}
\centering
\begin{threeparttable}[ht]
\footnotesize
\resizebox{0.99\linewidth}{!}{
\begin{tabular}{l|cc|cc|cc|cc}
\toprule
& \multicolumn{4}{c|}{Standard Set} & \multicolumn{4}{c}{Adversarial Set} \\
\cmidrule(lr){2-5} \cmidrule(lr){6-9}
& \multicolumn{2}{c|}{Naturalness} & \multicolumn{2}{c|}{Coherence} & \multicolumn{2}{c|}{Naturalness} & \multicolumn{2}{c}{Coherence} \\
Metrics & Pearson's $\rho$ & Spearman's $\tau$ & Pearson's $\rho$ & Spearman's $\tau$ & Pearson's $\rho$ & Spearman's $\tau$ & Pearson's $\rho$ & Spearman's $\tau$ \\
\midrule
BLEU-1 & 0.1793 & 0.1769 & 0.1924 & 0.1873 & 0.2012 & 0.2054 & 0.2138 & 0.2168 \\
BLEU-2 & 0.1836 & 0.1723 & 0.1947 & 0.1835 & 0.1487 & 0.1921 & 0.1598 & 0.2035 \\
BLEU-3 & 0.1802 & 0.1788 & 0.1906 & 0.1893 & 0.1496 & 0.1925 & 0.1599 & 0.2026 \\
BLEU-4 & 0.1932 & 0.1741 & 0.2053 & 0.1843 & 0.1549 & 0.1798 & 0.1648 & 0.1906 \\
ROUGE-1 & 0.2126 & 0.1976 & 0.2266 & 0.2087 & 0.2248 & 0.2088 & 0.2392 & 0.2207 \\
ROUGE-2 & 0.2001 & 0.1861 & 0.2134 & 0.1971 & 0.1462 & 0.1641 & 0.1574 & 0.1753 \\
ROUGE-L & 0.2129 & 0.1984 & 0.2258 & 0.2092 & 0.2318 & 0.2216 & 0.2459 & 0.2343 \\
METEOR & 0.1762 & 0.1521 & 0.1866 & 0.1612 & 0.1296 & 0.1002 & 0.1392 & 0.1071 \\
BERTScore & 0.2452 & 0.2593 & 0.2607 & 0.2742 & 0.1959 & 0.2176 & 0.2084 & 0.2301 \\
\midrule
DEB & 0.3152 & 0.2778 & 0.3356 & 0.2954 & 0.3402 & 0.3321 & 0.3617 & 0.3517 \\
USR & 0.2568 & 0.2419 & 0.2730 & 0.2567 & 0.2234 & 0.2693 & 0.2365 & 0.2845 \\
Mask-and-fill & 0.1852 & 0.2007 & 0.1969 & 0.2124 & 0.2532 & 0.2827 & 0.2684 & 0.2986 \\
MDD-Eval & 0.2739 & 0.2354 & 0.2918 & 0.2505 & 0.2907 & 0.2723 & 0.3083 & 0.2886 \\
\midrule
G-Eval (\texttt{GPT-3.5}) & 0.3326 & 0.3243 & 0.3512 & 0.3406 & 0.3201 & 0.3326 & 0.3411 & 0.3523 \\
QwQ-32B & 0.3822 & 0.3793 & 0.4023 & 0.3962 & 0.3692 & 0.3774 & 0.3898 & 0.3963 \\
Qwen2.5-7B & 0.3595 & 0.3615 & 0.3782 & 0.3796 & 0.3462 & 0.3584 & 0.3672 & 0.3779 \\
G-Eval (\texttt{GPT-4}) & 0.4219 & 0.4210 & 0.4458 & 0.4435 & 0.4183 & 0.4421 & 0.4443 & 0.4654 \\
LLM-Eval (\texttt{GPT-3.5}) & 0.3459 & 0.3638 & 0.3662 & 0.3834 & 0.3405 & 0.3335 & 0.3613 & 0.3528 \\
LLM-Eval (\texttt{GPT-4}) & 0.4210 & 0.4520 & 0.4452 & 0.4751 & 0.4573 & 0.4415 & 0.4842 & 0.4662 \\
\midrule
Ours(\texttt{w/o LLM}) & 0.3486 & 0.3557 & 0.3686 & 0.3754 & 0.3398 & 0.3452 & 0.3612 & 0.3658 \\
Ours (\texttt{GPT-3.5 w/o AMR}) & 0.4474 & 0.4476 & 0.4725 & 0.4723 & 0.4504 & 0.4628 & 0.4765 & 0.4880 \\
Ours (\texttt{GPT-3.5 w/o SLM}) & 0.4663 & 0.4605 & 0.4922 & 0.4856 & 0.4767 & 0.4775 & 0.5055 & 0.5044 \\
Ours (\texttt{GPT-3.5}) & 0.4768 & 0.4751 & 0.5035 & 0.5006 & 0.4826 & 0.4793 & 0.5107 & 0.5057 \\
Ours (\texttt{GPT-4 w/o AMR}) & 0.5158 & 0.5258 & 0.5449 & 0.5545 & 0.5086 & 0.5384 & 0.5367 & 0.5685 \\
Ours (\texttt{GPT-4 w/o SLM}) & 0.5292 & 0.5559 & 0.5586 & 0.5862 & 0.5391 & 0.5079 & 0.5681 & 0.5361 \\
Ours (\texttt{GPT-4}) & \textbf{0.5550} & \textbf{0.5779} & \textbf{0.5863} & \textbf{0.6093} & \textbf{0.5485} & \textbf{0.5680} & \textbf{0.5801} & \textbf{0.5991} \\
\midrule
& \multicolumn{2}{c|}{Engagingness} & \multicolumn{2}{c|}{Groundedness} & \multicolumn{2}{c|}{Engagingness} & \multicolumn{2}{c}{Groundedness} \\
Metrics & Pearson's $\rho$ & Spearman's $\tau$ & Pearson's $\rho$ & Spearman's $\tau$ & Pearson's $\rho$ & Spearman's $\tau$ & Pearson's $\rho$ & Spearman's $\tau$ \\
\midrule
BLEU-1 & 0.1824 & 0.1809 & 0.1823 & 0.1849 & 0.2059 & 0.2091 & 0.2087 & 0.2095 \\
BLEU-2 & 0.1868 & 0.1759 & 0.1873 & 0.1773 & 0.1527 & 0.1963 & 0.1548 & 0.1967 \\
BLEU-3 & 0.1832 & 0.1825 & 0.1848 & 0.1834 & 0.1537 & 0.1968 & 0.1540 & 0.1973 \\
BLEU-4 & 0.1967 & 0.1777 & 0.1968 & 0.1786 & 0.1593 & 0.1844 & 0.1602 & 0.1828 \\
ROUGE-1 & 0.2171 & 0.2026 & 0.2169 & 0.2026 & 0.2294 & 0.2141 & 0.2281 & 0.2135 \\
ROUGE-2 & 0.2042 & 0.1911 & 0.2043 & 0.1911 & 0.1501 & 0.1693 & 0.1502 & 0.1685 \\
ROUGE-L & 0.2176 & 0.2034 & 0.2183 & 0.2034 & 0.2367 & 0.2271 & 0.2364 & 0.2254 \\
METEOR & 0.1797 & 0.1561 & 0.1791 & 0.1561 & 0.1331 & 0.1034 & 0.1341 & 0.1059 \\
BERTScore & 0.2504 & 0.2658 & 0.2505 & 0.2639 & 0.2004 & 0.2230 & 0.2017 & 0.2221 \\
\midrule
DEB & 0.3212 & 0.2856 & 0.3224 & 0.2834 & 0.3480 & 0.3406 & 0.3469 & 0.3391 \\
USR & 0.2619 & 0.2482 & 0.2627 & 0.2460 & 0.2282 & 0.2760 & 0.2307 & 0.2742 \\
Mask-and-fill & 0.1895 & 0.2056 & 0.1900 & 0.2037 & 0.2592 & 0.2895 & 0.2608 & 0.2872 \\
MDD-Eval & 0.2798 & 0.2424 & 0.2797 & 0.2413 & 0.2975 & 0.2792 & 0.2963 & 0.2767 \\
\midrule
G-Eval (\texttt{GPT-3.5}) & 0.3352 & 0.3289 & 0.3483 & 0.3363 & 0.3273 & 0.3401 & 0.3291 & 0.3398 \\
QwQ-32B & 0.3844 & 0.3805 & 0.3972 & 0.3943 & 0.3757 & 0.3829 & 0.3826 & 0.3878 \\
Qwen2.5-7B & 0.3615 & 0.3625 & 0.3758 & 0.3774 & 0.3518 & 0.3635 & 0.3576 & 0.3698 \\
G-Eval (\texttt{GPT-4}) & 0.4264 & 0.4256 & 0.4342 & 0.4347 & 0.4264 & 0.4500 & 0.4302 & 0.4537 \\
LLM-Eval (\texttt{GPT-3.5}) & 0.3525 & 0.3723 & 0.3546 & 0.3697 & 0.3484 & 0.3421 & 0.3502 & 0.3379 \\
LLM-Eval (\texttt{GPT-4}) & 0.4283 & 0.4621 & 0.4315 & 0.4569 & 0.4660 & 0.4528 & 0.4689 & 0.4497 \\
\midrule
Ours(\texttt{w/o LLM}) & 0.3553 & 0.3646 & 0.3586 & 0.3625 & 0.3473 & 0.3545 & 0.3497 & 0.3524 \\
Ours (\texttt{GPT-3.5 w/o AMR}) & 0.4563 & 0.4592 & 0.4598 & 0.4577 & 0.4583 & 0.4745 & 0.4640 & 0.4727 \\
Ours (\texttt{GPT-3.5 w/o SLM}) & 0.4755 & 0.4723 & 0.4788 & 0.4708 & 0.4849 & 0.4902 & 0.4921 & 0.4887 \\
Ours (\texttt{GPT-3.5}) & 0.4865 & 0.4873 & 0.4892 & 0.4862 & 0.4906 & 0.4920 & 0.4981 & 0.4900 \\
Ours (\texttt{GPT-4 w/o AMR}) & 0.5263 & 0.5392 & 0.5299 & 0.5379 & 0.5171 & 0.5522 & 0.5230 & 0.5499 \\
Ours (\texttt{GPT-4 w/o SLM}) & 0.5398 & 0.5701 & 0.5429 & 0.5682 & 0.5481 & 0.5209 & 0.5531 & 0.5195 \\
Ours (\texttt{GPT-4}) & \textbf{0.5665} & \textbf{0.5927} & \textbf{0.5694} & \textbf{0.5909} & \textbf{0.5579} & \textbf{0.5826} & \textbf{0.5647} & \textbf{0.5807} \\
\bottomrule
\end{tabular}
}
\end{threeparttable}
\caption{Breakdown of Pearson and Spearman correlations with human judgments by evaluation criteria on the DailyDialog++ dataset.}
\label{tab:breakdown_dailydialog}
\end{table*}
\begin{table*}
\centering
\begin{threeparttable}[ht]
\footnotesize
\resizebox{0.99\linewidth}{!}{
\begin{tabular}{l|cc|cc|cc|cc}
\toprule
& \multicolumn{4}{c|}{Standard Set} & \multicolumn{4}{c}{Adversarial Set} \\
\cmidrule(lr){2-5} \cmidrule(lr){6-9}
& \multicolumn{2}{c|}{Naturalness} & \multicolumn{2}{c|}{Coherence} & \multicolumn{2}{c|}{Naturalness} & \multicolumn{2}{c}{Coherence} \\
Metrics & Pearson's $\rho$ & Spearman's $\tau$ & Pearson's $\rho$ & Spearman's $\tau$ & Pearson's $\rho$ & Spearman's $\tau$ & Pearson's $\rho$ & Spearman's $\tau$ \\
\midrule
BLEU-1 & 0.2014 & 0.2103 & 0.2115 & 0.2204 & 0.1725 & 0.1625 & 0.1807 & 0.1702 \\
BLEU-2 & 0.1905 & 0.1780 & 0.1996 & 0.1869 & 0.1369 & 0.1581 & 0.1443 & 0.1658 \\
BLEU-3 & 0.1638 & 0.1897 & 0.1722 & 0.1987 & 0.1340 & 0.1637 & 0.1412 & 0.1715 \\
BLEU-4 & 0.1954 & 0.1886 & 0.2052 & 0.1976 & 0.1220 & 0.1505 & 0.1286 & 0.1582 \\
ROUGE-1 & 0.2078 & 0.2106 & 0.2183 & 0.2214 & 0.2024 & 0.2145 & 0.2127 & 0.2251 \\
ROUGE-2 & 0.1967 & 0.1975 & 0.2065 & 0.2073 & 0.1786 & 0.2023 & 0.1879 & 0.2124 \\
ROUGE-L & 0.2051 & 0.1985 & 0.2155 & 0.2085 & 0.1978 & 0.2182 & 0.2077 & 0.2291 \\
METEOR & 0.1949 & 0.1725 & 0.2047 & 0.1812 & 0.1403 & 0.1667 & 0.1476 & 0.1744 \\
BERTScore & 0.2796 & 0.2656 & 0.2934 & 0.2788 & 0.2196 & 0.2579 & 0.2313 & 0.2708 \\
\midrule
DEB & 0.3562 & 0.3351 & 0.3744 & 0.3518 & 0.3424 & 0.3618 & 0.3601 & 0.3796 \\
USR & 0.3381 & 0.3370 & 0.3551 & 0.3542 & 0.3591 & 0.3765 & 0.3772 & 0.3954 \\
MDD-Eval & 0.3396 & 0.3328 & 0.3566 & 0.3492 & 0.3640 & 0.3513 & 0.3830 & 0.3689 \\
Mask-and-fill & 0.3017 & 0.3030 & 0.3168 & 0.3182 & 0.3673 & 0.3525 & 0.3856 & 0.3702 \\
\midrule
G-Eval (\texttt{GPT-3.5}) & 0.4773 & 0.4757 & 0.5009 & 0.4988 & 0.4442 & 0.4502 & 0.4662 & 0.4722 \\
QwQ-32B & 0.4905 & 0.4889 & 0.5148 & 0.5126 & 0.4662 & 0.4714 & 0.4894 & 0.4942 \\
Qwen2.5-7B & 0.4675 & 0.4617 & 0.4910 & 0.4844 & 0.4512 & 0.4598 & 0.4735 & 0.4819 \\
G-Eval (\texttt{GPT-4}) & 0.5115 & 0.5195 & 0.5368 & 0.5444 & 0.5002 & 0.5391 & 0.5246 & 0.5640 \\
LLM-Eval (\texttt{GPT-3.5}) & 0.4539 & 0.4464 & 0.4757 & 0.4683 & 0.4343 & 0.4506 & 0.4557 & 0.4726 \\
LLM-Eval (\texttt{GPT-4}) & 0.5193 & 0.5265 & 0.5449 & 0.5522 & 0.5142 & 0.5131 & 0.5396 & 0.5385 \\
\midrule
Ours(\texttt{w/o LLM}) & 0.3582 & 0.3696 & 0.3754 & 0.3873 & 0.3858 & 0.3815 & 0.4052 & 0.4007 \\
Ours (\texttt{GPT-3.5 w/o AMR}) & 0.4887 & 0.4878 & 0.5127 & 0.5119 & 0.4892 & 0.4982 & 0.5131 & 0.5229 \\
Ours (\texttt{GPT-3.5 w/o SLM}) & 0.4995 & 0.4946 & 0.5240 & 0.5192 & 0.5074 & 0.5062 & 0.5325 & 0.5313 \\
Ours(\texttt{GPT-3.5}) & 0.5385 & 0.5084 & 0.5648 & 0.5335 & 0.5079 & 0.5099 & 0.5329 & 0.5352 \\
Ours (\texttt{GPT-4 w/o AMR}) & 0.6051 & 0.5982 & 0.6347 & 0.6273 & 0.6029 & 0.5862 & 0.6327 & 0.6149 \\
Ours (\texttt{GPT-4 w/o SLM}) & 0.6116 & 0.6149 & 0.6418 & 0.6450 & 0.6094 & 0.6158 & 0.6397 & 0.6462 \\
Ours (\texttt{GPT-4}) & \textbf{0.6441} & \textbf{0.6448} & \textbf{0.6756} & \textbf{0.6762} & \textbf{0.6370} & \textbf{0.6458} & \textbf{0.6683} & \textbf{0.6768} \\
\midrule
& \multicolumn{2}{c|}{Engagingness} & \multicolumn{2}{c|}{Groundedness} & \multicolumn{2}{c|}{Engagingness} & \multicolumn{2}{c}{Groundedness} \\
Metrics & Pearson's $\rho$ & Spearman's $\tau$ & Pearson's $\rho$ & Spearman's $\tau$ & Pearson's $\rho$ & Spearman's $\tau$ & Pearson's $\rho$ & Spearman's $\tau$ \\
\midrule
BLEU-1 & 0.2052 & 0.2142 & 0.2072 & 0.2158 & 0.1755 & 0.1657 & 0.1769 & 0.1669 \\
BLEU-2 & 0.1940 & 0.1813 & 0.1961 & 0.1830 & 0.1397 & 0.1616 & 0.1410 & 0.1621 \\
BLEU-3 & 0.1670 & 0.1933 & 0.1688 & 0.1947 & 0.1367 & 0.1673 & 0.1380 & 0.1680 \\
BLEU-4 & 0.1992 & 0.1921 & 0.2010 & 0.1936 & 0.1244 & 0.1537 & 0.1260 & 0.1548 \\
ROUGE-1 & 0.2119 & 0.2147 & 0.2139 & 0.2168 & 0.2067 & 0.2194 & 0.2082 & 0.2203 \\
ROUGE-2 & 0.2004 & 0.2014 & 0.2027 & 0.2030 & 0.1823 & 0.2066 & 0.1839 & 0.2079 \\
ROUGE-L & 0.2091 & 0.2024 & 0.2116 & 0.2043 & 0.2017 & 0.2225 & 0.2036 & 0.2244 \\
METEOR & 0.1987 & 0.1758 & 0.2005 & 0.1777 & 0.1430 & 0.1697 & 0.1446 & 0.1711 \\
BERTScore & 0.2850 & 0.2707 & 0.2881 & 0.2733 & 0.2240 & 0.2633 & 0.2265 & 0.2652 \\
\midrule
DEB & 0.3626 & 0.3417 & 0.3678 & 0.3452 & 0.3489 & 0.3687 & 0.3533 & 0.3722 \\
USR & 0.3444 & 0.3434 & 0.3488 & 0.3478 & 0.3663 & 0.3838 & 0.3698 & 0.3877 \\
MDD-Eval & 0.3460 & 0.3392 & 0.3503 & 0.3428 & 0.3712 & 0.3581 & 0.3760 & 0.3621 \\
Mask-and-fill & 0.3073 & 0.3087 & 0.3114 & 0.3123 & 0.3744 & 0.3594 & 0.3781 & 0.3631 \\
\midrule
G-Eval (\texttt{GPT-3.5}) & 0.4862 & 0.4847 & 0.4928 & 0.4904 & 0.4527 & 0.4587 & 0.4573 & 0.4629 \\
QwQ-32B & 0.4997 & 0.4981 & 0.5058 & 0.5038 & 0.4749 & 0.4801 & 0.4809 & 0.4852 \\
Qwen2.5-7B & 0.4762 & 0.4704 & 0.4822 & 0.4761 & 0.4598 & 0.4683 & 0.4645 & 0.4728 \\
G-Eval (\texttt{GPT-4}) & 0.5209 & 0.5290 & 0.5271 & 0.5342 & 0.5098 & 0.5489 & 0.5146 & 0.5532 \\
LLM-Eval (\texttt{GPT-3.5}) & 0.4624 & 0.4548 & 0.4673 & 0.4597 & 0.4426 & 0.4590 & 0.4474 & 0.4639 \\
LLM-Eval (\texttt{GPT-4}) & 0.5284 & 0.5357 & 0.5356 & 0.5425 & 0.5234 & 0.5223 & 0.5304 & 0.5292 \\
\midrule
Ours(\texttt{w/o LLM}) & 0.3651 & 0.3767 & 0.3685 & 0.3798 & 0.3934 & 0.3890 & 0.3973 & 0.3930 \\
Ours (\texttt{GPT-3.5 w/o AMR}) & 0.4978 & 0.4969 & 0.5037 & 0.5027 & 0.4984 & 0.5075 & 0.5037 & 0.5131 \\
Ours (\texttt{GPT-3.5 w/o SLM}) & 0.5089 & 0.5039 & 0.5150 & 0.5097 & 0.5170 & 0.5158 & 0.5226 & 0.5216 \\
Ours(\texttt{GPT-3.5}) & 0.5481 & 0.5175 & 0.5553 & 0.5241 & 0.5173 & 0.5194 & 0.5234 & 0.5253 \\
Ours (\texttt{GPT-4 w/o AMR}) & 0.6158 & 0.6088 & 0.6240 & 0.6166 & 0.6139 & 0.5969 & 0.6217 & 0.6038 \\
Ours (\texttt{GPT-4 w/o SLM}) & 0.6227 & 0.6261 & 0.6307 & 0.6334 & 0.6206 & 0.6269 & 0.6283 & 0.6347 \\
Ours (\texttt{GPT-4}) & \textbf{0.6555} & \textbf{0.6562} & \textbf{0.6639} & \textbf{0.6643} & \textbf{0.6483} & \textbf{0.6571} & \textbf{0.6567} & \textbf{0.6650} \\
\bottomrule
\end{tabular}
}
\end{threeparttable}
\caption{Breakdown of Pearson and Spearman correlations with human judgments by evaluation criteria on the PersonaChat dataset.}
\label{tab:breakdown_personachat}
\end{table*}

\begin{table*}
\centering
\begin{threeparttable}[ht]
\footnotesize
\resizebox{0.99\linewidth}{!}{
\begin{tabular}{l|cc|cc|cc|cc}
\toprule
& \multicolumn{4}{c|}{Standard Set} & \multicolumn{4}{c}{Adversarial Set} \\
\cmidrule(lr){2-5} \cmidrule(lr){6-9}
& \multicolumn{2}{c|}{Naturalness} & \multicolumn{2}{c|}{Coherence} & \multicolumn{2}{c|}{Naturalness} & \multicolumn{2}{c}{Coherence} \\
Metrics & Pearson's $\rho$ & Spearman's $\tau$ & Pearson's $\rho$ & Spearman's $\tau$ & Pearson's $\rho$ & Spearman's $\tau$ & Pearson's $\rho$ & Spearman's $\tau$ \\
\midrule
BLEU-1 & 0.2055 & 0.1938 & 0.2149 & 0.2026 & 0.1408 & 0.1517 & 0.1480 & 0.1588 \\
BLEU-2 & 0.1682 & 0.1732 & 0.1759 & 0.1812 & 0.1261 & 0.1403 & 0.1330 & 0.1475 \\
BLEU-3 & 0.1541 & 0.1605 & 0.1613 & 0.1679 & 0.1191 & 0.1293 & 0.1259 & 0.1363 \\
BLEU-4 & 0.1449 & 0.1468 & 0.1515 & 0.1538 & 0.1289 & 0.1198 & 0.1358 & 0.1258 \\
ROUGE-1 & 0.2004 & 0.2095 & 0.2096 & 0.2193 & 0.1710 & 0.1747 & 0.1793 & 0.1829 \\
ROUGE-2 & 0.1958 & 0.1982 & 0.2049 & 0.2074 & 0.1789 & 0.1983 & 0.1880 & 0.2075 \\
ROUGE-L & 0.2148 & 0.1965 & 0.2246 & 0.2056 & 0.1863 & 0.2283 & 0.1953 & 0.2388 \\
METEOR & 0.1815 & 0.1539 & 0.1899 & 0.1612 & 0.1483 & 0.1647 & 0.1554 & 0.1724 \\
BERTScore & 0.2497 & 0.2485 & 0.2613 & 0.2600 & 0.2142 & 0.2501 & 0.2245 & 0.2617 \\
\midrule
DEB & 0.3181 & 0.3230 & 0.3330 & 0.3381 & 0.3341 & 0.3585 & 0.3497 & 0.3752 \\
USR & 0.3387 & 0.3350 & 0.3545 & 0.3507 & 0.3258 & 0.1667 & 0.3419 & 0.1745 \\
MDD-Eval & 0.3201 & 0.3320 & 0.3353 & 0.3476 & 0.3777 & 0.3475 & 0.3961 & 0.3638 \\
Mask-and-fill & 0.2929 & 0.2982 & 0.3068 & 0.3123 & 0.3582 & 0.3545 & 0.3754 & 0.3710 \\
\midrule
G-Eval (\texttt{GPT-3.5}) & 0.4880 & 0.4645 & 0.5108 & 0.4866 & 0.4663 & 0.4578 & 0.4883 & 0.4798 \\
QwQ-32B & 0.4975 & 0.4724 & 0.5208 & 0.4950 & 0.4773 & 0.4714 & 0.5004 & 0.4936 \\
Qwen2.5-7B & 0.4814 & 0.4596 & 0.5041 & 0.4810 & 0.4591 & 0.4569 & 0.4812 & 0.4788 \\
G-Eval (\texttt{GPT-4}) & 0.5192 & 0.4939 & 0.5434 & 0.5173 & 0.4880 & 0.4907 & 0.5107 & 0.5137 \\
LLM-Eval (\texttt{GPT-3.5}) & 0.4724 & 0.4687 & 0.4949 & 0.4909 & 0.4408 & 0.4688 & 0.4618 & 0.4909 \\
LLM-Eval (\texttt{GPT-4}) & 0.4893 & 0.4979 & 0.5123 & 0.5214 & 0.5059 & 0.5137 & 0.5298 & 0.5378 \\
\midrule
Ours(\texttt{w/o LLM}) & 0.3520 & 0.3516 & 0.3685 & 0.3681 & 0.3529 & 0.3505 & 0.3694 & 0.3669 \\
Ours (\texttt{GPT-3.5 w/o AMR}) & 0.4907 & 0.5003 & 0.5138 & 0.5237 & 0.5001 & 0.4981 & 0.5235 & 0.5216 \\
Ours (\texttt{GPT-3.5 w/o SLM}) & 0.5053 & 0.4981 & 0.5290 & 0.5217 & 0.4994 & 0.4983 & 0.5228 & 0.5218 \\
Ours(\texttt{GPT-3.5}) & 0.5080 & 0.4997 & 0.5319 & 0.5233 & 0.5009 & 0.4993 & 0.5245 & 0.5228 \\
Ours (\texttt{GPT-4 w/o AMR}) & 0.6130 & 0.6122 & 0.6417 & 0.6410 & 0.6055 & 0.5087 & 0.6338 & 0.5326 \\
Ours (\texttt{GPT-4 w/o SLM}) & 0.6325 & 0.6336 & 0.6615 & 0.6629 & 0.6251 & 0.6258 & 0.6546 & 0.6546 \\
Ours (\texttt{GPT-4}) & \textbf{0.6490} & \textbf{0.6455} & \textbf{0.6789} & \textbf{0.6752} & \textbf{0.6449} & \textbf{0.6524} & \textbf{0.6748} & \textbf{0.6822} \\
\midrule
& \multicolumn{2}{c|}{Engagingness} & \multicolumn{2}{c|}{Groundedness} & \multicolumn{2}{c|}{Engagingness} & \multicolumn{2}{c}{Groundedness} \\
Metrics & Pearson's $\rho$ & Spearman's $\tau$ & Pearson's $\rho$ & Spearman's $\tau$ & Pearson's $\rho$ & Spearman's $\tau$ & Pearson's $\rho$ & Spearman's $\tau$ \\
\midrule
BLEU-1 & 0.2091 & 0.1971 & 0.2111 & 0.1984 & 0.1435 & 0.1546 & 0.1452 & 0.1561 \\
BLEU-2 & 0.1714 & 0.1765 & 0.1731 & 0.1777 & 0.1287 & 0.1429 & 0.1304 & 0.1449 \\
BLEU-3 & 0.1569 & 0.1635 & 0.1585 & 0.1647 & 0.1215 & 0.1318 & 0.1231 & 0.1339 \\
BLEU-4 & 0.1474 & 0.1494 & 0.1490 & 0.1510 & 0.1312 & 0.1220 & 0.1333 & 0.1235 \\
ROUGE-1 & 0.2037 & 0.2130 & 0.2065 & 0.2151 & 0.1742 & 0.1780 & 0.1764 & 0.1796 \\
ROUGE-2 & 0.1993 & 0.2019 & 0.2015 & 0.2035 & 0.1824 & 0.2020 & 0.1847 & 0.2038 \\
ROUGE-L & 0.2186 & 0.2000 & 0.2207 & 0.2023 & 0.1899 & 0.2325 & 0.1922 & 0.2344 \\
METEOR & 0.1845 & 0.1567 & 0.1868 & 0.1575 & 0.1509 & 0.1675 & 0.1527 & 0.1693 \\
BERTScore & 0.2542 & 0.2530 & 0.2568 & 0.2553 & 0.2183 & 0.2547 & 0.2207 & 0.2575 \\
\midrule
DEB & 0.3237 & 0.3288 & 0.3273 & 0.3324 & 0.3402 & 0.3649 & 0.3439 & 0.3689 \\
USR & 0.3448 & 0.3409 & 0.3487 & 0.3448 & 0.3320 & 0.1698 & 0.3358 & 0.1715 \\
MDD-Eval & 0.3259 & 0.3380 & 0.3296 & 0.3415 & 0.3847 & 0.3539 & 0.3892 & 0.3576 \\
Mask-and-fill & 0.2983 & 0.3037 & 0.3012 & 0.3067 & 0.3649 & 0.3609 & 0.3688 & 0.3646 \\
\midrule
G-Eval (\texttt{GPT-3.5}) & 0.4970 & 0.4728 & 0.5033 & 0.4775 & 0.4750 & 0.4665 & 0.4800 & 0.4710 \\
QwQ-32B & 0.5067 & 0.4809 & 0.5127 & 0.4852 & 0.4863 & 0.4801 & 0.4909 & 0.4846 \\
Qwen2.5-7B & 0.4902 & 0.4681 & 0.4962 & 0.4731 & 0.4678 & 0.4652 & 0.4727 & 0.4694 \\
G-Eval (\texttt{GPT-4}) & 0.5286 & 0.5027 & 0.5345 & 0.5080 & 0.4970 & 0.4996 & 0.5022 & 0.5048 \\
LLM-Eval (\texttt{GPT-3.5}) & 0.4811 & 0.4773 & 0.4864 & 0.4823 & 0.4489 & 0.4772 & 0.4532 & 0.4826 \\
LLM-Eval (\texttt{GPT-4}) & 0.4981 & 0.5069 & 0.5036 & 0.5125 & 0.5147 & 0.5226 & 0.5207 & 0.5287 \\
\midrule
Ours(\texttt{w/o LLM}) & 0.3584 & 0.3580 & 0.3621 & 0.3616 & 0.3593 & 0.3569 & 0.3629 & 0.3605 \\
Ours (\texttt{GPT-3.5 w/o AMR}) & 0.4997 & 0.5094 & 0.5046 & 0.5145 & 0.5093 & 0.5074 & 0.5143 & 0.5124 \\
Ours (\texttt{GPT-3.5 w/o SLM}) & 0.5146 & 0.5073 & 0.5200 & 0.5124 & 0.5089 & 0.5076 & 0.5136 & 0.5127 \\
Ours(\texttt{GPT-3.5}) & 0.5173 & 0.5089 & 0.5226 & 0.5142 & 0.5102 & 0.5087 & 0.5153 & 0.5132 \\
Ours (\texttt{GPT-4 w/o AMR}) & 0.6247 & 0.6239 & 0.6304 & 0.6293 & 0.6169 & 0.5173 & 0.6229 & 0.5242 \\
Ours (\texttt{GPT-4 w/o SLM}) & 0.6444 & 0.6456 & 0.6506 & 0.6517 & 0.6366 & 0.6372 & 0.6429 & 0.6430 \\
Ours (\texttt{GPT-4}) & \textbf{0.6612} & \textbf{0.6576} & \textbf{0.6674} & \textbf{0.6638} & \textbf{0.6566} & \textbf{0.6641} & \textbf{0.6631} & \textbf{0.6709} \\
\bottomrule
\end{tabular}
}
\end{threeparttable}
\caption{Breakdown of Pearson and Spearman correlations with human judgments by evaluation criteria on the TopicalChat dataset.}
\label{tab:breakdown_topicalchat}
\end{table*}
\section{Performance Breakdown by Evaluation Criteria}
\label{app:breakdown}
Tables~\ref{tab:breakdown_dailydialog}, \ref{tab:breakdown_personachat}, and \ref{tab:breakdown_topicalchat} present the correlation results broken down by individual evaluation criteria (Naturalness, Coherence, Engagingness, and Groundedness) for each dataset. This detailed analysis reveals that our method consistently outperforms baselines across all criteria, with particularly notable improvements in Coherence and Groundedness for adversarial examples. This pattern aligns with our expectation that AMR graph information would be especially beneficial for capturing semantic inconsistencies that affect contextual appropriateness.

\section{Example Prompt with AMR Graph and SLM Score}
\label{app:prompt_example}
In Table~\autoref{table:example}, we provide a concrete example of how the SLM score and AMR graph information are incorporated into the LLM prompt for evaluation.
\begin{table*}[b]
    \footnotesize
    \centering
\begin{tabular}{@{}p{\linewidth}@{}}
\toprule

\textbf{Prompt for Dialogue Response Evaluation} \\
\midrule
Rate the dialogue response.\\
\\
Use the prediction probability from the SLMs and AMR graphs of the conversation pair to aid your judgment.\\
\\
Note: Please take the time to fully read and understand the dialogue response.\\
\\
How coherent is the text of the dialogue response? (on a scale of 1-5, with 1 being the lowest)\\
\\
\textbf{Input:}\\
Conversation Context: Would you recommend some places for sightseeing? How about great canyon? Is it worth seeing?\\
\\
Response: The movie was really good, it was worth watching it.\\
\\
\textbf{AMR Graph:}\\
(multi-sentence\\
\ \ :snt1 (recommend\\
\ \ \ \ :ARG0 (you)\\
\ \ \ \ :polarity (amr-unk)\\
\ \ \ \ :ARG1 (place\\
\ \ \ \ \ \ :quant (some)\\
\ \ \ \ \ \ :location (sightsee)))\\
\ \ :snt2 (canyon\\
\ \ \ \ :mod (great)\\
\ \ \ \ :polarity (amr-unk))\\
\ \ \ \ :ARG1 (worth)\\
\ \ \ \ \ \ :ARG2 (see)\\
\ \ :snt3 (and\\
\ \ \ \ :mod (worth\\
\ \ \ \ \ \ :ARG1 (watch)\\
\ \ \ \ \ \ \ \ :ARG1 (movie)\\
\ \ \ \ :mod (good\\
\ \ \ \ \ \ :ARG1 (movie)))))\\
\\
\textbf{SLM score:} 0.32\\
\\
\textbf{Evaluation Form (Score ONLY):}\\
Coherence: \\
\bottomrule
\end{tabular}
    \caption{Example of prompt template showing how SLM score and AMR graph information are integrated to evaluate dialogue response coherence.}
    \label{tab:prompt_example}
\end{table*}
\label{table:example}
\end{document}